\documentclass[
]{ceurart}

\usepackage[utf8]{inputenc}
\usepackage{listings}

\usepackage[most]{tcolorbox}
\begin{document}

\copyrightyear{2025}
\copyrightclause{Copyright for this paper by its authors.
  Use permitted under Creative Commons License Attribution 4.0
  International (CC BY 4.0).}

\conference{CLEF 2025 Working Notes, September 9 -- 12 September 2025, Madrid, Spain}

\title{XplaiNLP at CheckThat! 2025: Multilingual Subjectivity Detection with Finetuned Transformers and Prompt-Based Inference with Large Language Models}

\title[mode=sub]{Notebook for the CheckThat! Lab at CLEF 2025}

\author[1,2]{Ariana Sahitaj}[%
orcid=0009-0002-0096-9383,
email=ariana.sahitaj@campus.tu-berlin.de,
]
\cormark[1]

\author[1,2]{Jiaao Li}[%
email=jiaao.li@campus.tu-berlin.de,
]

\author[1,2]{Pia {Wenzel Neves}}[%
orcid=0009-0001-8340-0215,
email=p.wenzel.2@campus.tu-berlin.de,
]

\author[1,2]{Fedor Splitt}[%
email=splitt@campus.tu-berlin.de,
]

\author[1,2]{Premtim Sahitaj}[%
orcid=0000-0003-3908-5681,
email=sahitaj@tu-berlin.de,
]

\author[1,2]{Charlott Jakob}[%
orcid=0009-0002-6262-9018,
email=c.jakob@tu-berlin.de,
]

\author[1,2]{Veronika Solopova}[
orcid=0000-0003-0183-9433,
email=veronika.solopova@tu-berlin.de,
]

\author[1,2]{Vera Schmitt}[%
orcid=0000-0002-9735-6956,
email=vera.schmitt@tu-berlin.de,
]

\address[1]{Quality and Usability Lab, Technische Universität Berlin, Berlin, Germany}
\address[2]{German Research Center for Artificial Intelligence (DFKI), Berlin, Germany}

\cortext[1]{Corresponding author.}






\begin{abstract}
This notebook reports the XplaiNLP submission to the CheckThat! 2025 shared task \cite{10.1007/978-3-031-88720-8_68} on multilingual subjectivity detection. We evaluate two approaches: (1) supervised fine-tuning of transformer encoders, EuroBERT, XLM-RoBERTa, and German-BERT, on monolingual and machine-translated training data; and (2) zero-shot prompting using two LLMs: \texttt{o3-mini} for Annotation (rule-based labelling) and \texttt{gpt-4.1-mini} for DoubleDown (contrastive rewriting) and Perspective (comparative reasoning). The Annotation Approach achieves 1\textsuperscript{st} place in the Italian monolingual subtask with an F$_1$ score of 0.8104, outperforming the baseline of 0.6941. In the Romanian zero-shot setting, the fine-tuned XLM-RoBERTa model obtains an F$_1$ score of 0.7917, ranking 3\textsuperscript{rd} and exceeding the baseline of 0.6461. The same model also performs reliably in the multilingual task and improves over the baseline in Greek. For German, a German-BERT model fine-tuned on translated training data from typologically related languages yields competitive performance over the baseline. In contrast, performance in the Ukrainian and Polish zero-shot settings falls slightly below the respective baselines, reflecting the challenge of generalization in low-resource cross-lingual scenarios.

\end{abstract}


\maketitle

\section{Introduction}

Understanding whether a sentence expresses a personal opinion or presents information in a neutral and therefore objective way is important in many natural language processing tasks \cite{yu2003towards}. This distinction is particularly relevant in the context of news reporting, where objectivity is traditionally considered a core principle. Yet, subjective or evaluative language is often embedded in news texts through stylistic choices and subtle dialogic elements that influence how readers interpret information. \cite{ruotsalainen2021future} This effect is especially strong when opinionated language is presented in the style of factual reporting, causing evaluative statements to appear as objective observations \cite{rodrigo2024systematic}. A precise distinction between subjective and objective language is important for tasks such as sentiment analysis \cite{chaturvedi2018distinguishing}, stance detection \cite{kasnesis2021combating}, automated fact-checking \cite{guo2022survey, sahitaj2025towards}, propaganda detection \cite{scheffler2021telegram}, argument mining \cite{sahitaj2024construction}, and bias identification \cite{riloff2003learning}. These applications rely on the ability to detect whether a statement reflects personal opinion, emotional language, or evaluative framing, or whether it is intended to convey factual content. Subjective sentences commonly include emotional terms, value judgments, or rhetorical elements such as irony or exaggeration \cite{palshikar2016learning}. However, even for human readers, it is not always simple to decide whether a sentence is subjective or not. Interpretations often depend on context and background knowledge, making critical thinking essential for distinguishing between evaluative language and factual reporting \cite{liu2010sentiment, sahitaj2024towards}. This issue becomes more apparent in multilingual settings, as different languages signal subjectivity in diverse ways, through verb forms, word order, lexical choices, or stylistic conventions \cite{eid2025bridging}. At the same time, many languages lack annotated resources for subjectivity detection, which poses an additional challenge for training reliable models \cite{kocon2021learning}.\\
This notebook describes our submission to the CheckThat! Lab at CLEF 2025 \cite{clef-checkthat:2025:task1}, which focused on sentence-level subjectivity detection across multiple languages. Our approach investigates two complementary approaches: (1) supervised fine-tuning of multilingual and monolingual transformer-based encoders on annotated datasets, and (2) zero-shot prompting with LLMs using natural language inference guided by explicit instructions. While the first approach relies on parameter-efficient adaptation of pretrained models, the second uses the contextual reasoning capabilities of LLMs to classify subjectivity without additional training.\\
The paper is structured as follows: In Section~\ref{sec:related-work} we review related work on subjectivity detection. Section~\ref{sec:dataset} introduces the dataset used in the shared task. Section~\ref{sec:approach} details our modeling approaches, including fine-tuned transformer models and zero-shot prompting strategies. Section~\ref{sec:experiments} presents our evaluation results before we finally summarize our findings in Section~\ref{sec:conclusion} and outline directions for future work.

\section{Related Work}
\label{sec:related-work}


The aim of subjectivity detection is to distinguish language that conveys private states, such as opinions, from language that presents information in a way that seem factual or in a neutral manner, regardless of whether the information is actually true \cite{wiebe2004learning}. \citeauthor{yu2003towards} proposed one of the earliest computational models for sentence-level subjectivity detection, using Bayesian classification. \citeauthor{riloff2003learning} and \citeauthor{riloff2005exploiting} explored how subjectivity detection can improve information extraction by reducing false positives, especially in metaphorical or emotional contexts. These studies marked a shift toward integrating subjectivity classification into practical NLP pipelines. In 2006 \citeauthor{esuli2006sentiwordnet} introduced SentiWordNet, a lexical resource assigning polarity and objectivity scores to WordNet synsets. While primarily intended for polarity detection, it also provides objectivity measures, implicitly supporting subjectivity detection tasks. \citeauthor{banea2008bootstrapping} addressed the scarcity of resources in non-English languages by developing a bootstrapping approach to build subjectivity lexicons using only seed lists and a basic corpus. This method made subjectivity analysis feasible for low-resource settings. Later, \citeauthor{chaturvedi2018distinguishing} provided a comprehensive survey of both rule-based and automatic models. They emphasized that subjectivity detection is a important preprocessing step for sentiment analysis, as polarity classifiers may otherwise incorrectly label neutral statements as emotional content. \citeauthor{antici2023corpus} introduced NewsSD-ENG, a sentence-level dataset with human-annotation subjectivity labels. Their experiments showed that multilingual transformer models (M-BERT and M-SBERT) clearly outperformed traditional classifiers (SVM and LR), with multilingual training improving performance and enabling robust cross-lingual subjectivity detection. \citeauthor{savinova2023analyzing} reframed subjectivity detection as a regression task and fine-tuned a RoBERTa model to predict degrees of subjectivity in news texts. Their model aligned closely with human judgments and significantly outperformed a widely used rule-based regressor based on lexical patterns. Recent work by \citeauthor{shokri2024subjectivity} evaluated large language models like GPT-3.5, GPT-4, and Gemini for sentence-level subjectivity detection in English news. They showed that while fine-tuned models perform well in-distribution, zero-shot and chain-of-thought prompting yield more robust generalization across diverse datasets. Also, \citeauthor{suwaileh2024thatiar} introduced ThatiAR, the first large-scale dataset for sentence-level subjectivity detection in Arabic news. They demonstrated that GPT-4, especially in few-shot settings, outperformed traditional and fine-tuned Arabic language models, highlighting the potential of LLMs in morphologically-rich and culturally diverse contexts.

\section{Dataset}
\label{sec:dataset}

The dataset employed in this study originates from the shared task on subjectivity detection from the CheckThat! Lab 2025.\footnote{https://checkthat.gitlab.io/clef2025/task1/} It is designed to evaluate the ability of computational systems to classify whether a sentence or short paragraph extracted from a news article expresses a subjective (\textbf{SUBJ}) or objective opinion (\textbf{OBJ}). The dataset comprises textual instances drawn from news sources in five languages: English, Italian, German, Bulgarian, and Arabic. For each language, the data is divided into four subsets: training, development, development-test, and test (with test labels withheld for evaluation purposes). 
\begin{table}[h]
  \caption{Sentence and label distribution per language and dataset split. Each split shows total number of sentences and class distribution.}
  \label{tab:dataset-stats}
  \centering
  \begin{tabular}{@{}lccc ccc ccc@{}}
    \toprule
    \textbf{Language} 
    & \multicolumn{3}{c}{\textbf{Train}} 
    & \multicolumn{3}{c}{\textbf{Dev}} 
    & \multicolumn{3}{c}{\textbf{Dev-Test}} \\
    \cmidrule(lr){2-4} \cmidrule(lr){5-7} \cmidrule(lr){8-10}
    & Total & SUBJ & OBJ & Total & SUBJ & OBJ & Total & SUBJ & OBJ \\
    \midrule
    English   & 830  & 298 & 532 
             & 462  & 240 & 222 
             & 484  & 122 & 362 \\
    Italian   & 1613 & 382 & 1231 
             & 667  & 177 & 490 
             & 513  & 136 & 377 \\
    German    & 800  & 308 & 492 
             & 491  & 174 & 317 
             & 337  & 111 & 226 \\
    Bulgarian & 729  & 323 & 406 
             & 467  & 292 & 175 
             & 250  & 107 & 143 \\
    Arabic    & 2446 & 1055 & 1391 
             & 742  & 476 & 266 
             & 748  & 323  & 425 \\
    \bottomrule
  \end{tabular}
\end{table}
Table~\ref{tab:dataset-stats} provides a detailed overview of the sentence and label distribution across languages and dataset splits. Among the five languages, Arabic stands out with the largest training set (2,446 instances), while the remaining languages, Italian (1,613), English (830), German (800), and Bulgarian (729), are comparatively balanced in size. These differences likely reflect variation in source availability, annotation resources, and data curation priorities. The distribution of labels also varies by language and split: In the English training set, 64.10\% of sentences are labeled as objective, while the remainder are subjective. The Italian data exhibits a strong bias towards objectivity across all splits, with over 73\% of sentences labeled as OBJ. By contrast, the Bulgarian development set features a higher proportion of subjective content, highlighting potential cultural or editorial differences in reporting styles. 
\begin{table}[h]
  \caption{Token statistics for each language on the test set.}
  \label{tab:testset-stats}
  \centering
  \begin{tabular}{lccccc}
    \toprule
    \multicolumn{6}{c}{\textbf{Test Set}} \\
    \midrule
    \textbf{Language} 
    & \textbf{Total} & \textbf{Avg. Length} & \textbf{Min} & \textbf{Max} & \textbf{Median} \\
    \midrule
    English             & 300  & 28.79 & 2  & 114 & 28.0 \\
    Italian             & 299  & 28.34 & 2  & 113 & 23.0 \\
    German              & 347  & 31.46 & 8  & 117 & 28.0 \\
    Arabic              & 1036 & 42.39 & 5  & 175 & 38.0 \\
    \midrule
    Zero-Shot Romanian  & 206  & 34.56 & 2  & 151 & 30.0 \\
    Zero-Shot Ukrainian & 297  & 28.17 & 2  & 114 & 26.0 \\
    Zero-Shot Greek     & 284  & 40.88 & 1  & 141 & 36.0 \\
    Zero-Shot Polish    & 351  & 29.43 & 4  & 97  & 26.0 \\
    \bottomrule
  \end{tabular}
\end{table}

Token-level statistics for the test set are summarized in Table~\ref{tab:testset-stats}, using the \texttt{xlm-roberta-base} tokenizer \cite{xlm-roberta}. Arabic and Greek test sets have the longest sequences on average, while English, Italian, and Polish are more concise. These differences may impact model robustness across languages. During preprocessing, we identified several anomalous cases, particularly in the Bulgarian, German, and Italian splits, where open-ended quotation marks led to excessively long token sequences (often over 500 tokens). These likely stemmed from the tokenizer's handling of unmatched punctuation and were manually corrected to avoid distortions in length statistics.

\subsection{Ambiguities and Hard-to-Translate Cases}
To better understand common sources of labeling disagreement, we manually examined 20 sentences with annotation conflicts in the English development set. Five recurring themes emerged, each highlighting linguistic or contextual features that challenge binary subjectivity classification:

\begin{itemize}
    \item \textbf{Immigration Discourse:} Statements such as \textit{“Mr. Buchanan's criticism of immigration”} may appear factual but are often ideologically charged, subtly framing the topic in ways that evoke subjective interpretation.
    
    \item \textbf{Race and Social Commentary:} Sentences referencing phrases such as \textit{“CRT anti-white curricula”} or \textit{“diversity, equity, and inclusion”} are lexically neutral but semantically charged. The underlying ideological associations can trigger differing interpretations, often reflecting the annotator’s sociopolitical context.

    \item \textbf{Media and Political Rhetoric:} Labels such as \textit{“Lügenpresse”} or \textit{“Treason Lobby”} embed explicit bias or contempt within declarative syntax, complicating detection by surface-level classifiers.

    \item \textbf{Sarcasm and Pragmatic Devices:} Utterances like \textit{“What could possibly go wrong?”} rely on irony or context-based inference. Lacking overt opinion markers, they remain difficult to detect using standard lexical cues.

    \item \textbf{Framing in Economic and Environmental Topics:} Sentences such as \textit{“bribed by a globalist billionaire”} combine factual assertions with emotionally charged language, blurring the line between reporting and commentary.
\end{itemize}

\section{Approach}
\label{sec:approach}
We explore two complementary approaches to multilingual subjectivity detection: supervised fine-tuning of transformer-based classifiers and zero-shot prompting with LLMs. The former trains task-specific classifiers on available annotated data, while the latter uses instruction-following capabilities of LLMs to perform inference without parameter updates. This section details the setup, training procedures, and reasoning strategies employed in both directions.

\subsection{Fine-Tuned Transformers}

\paragraph{Fine-tuning German-BERT with Translated Training Data}

The idea was to enlarge the German training dataset by translating the other provided training datasets into German and fine-tuning a BERT model \cite{bert-base-german-cased}. We ordered the languages from most to least similar to German, assuming that using the translation of training data of more similar languages will yield better results. First English (West Germanic, Indo-European), then Italian (Romance, Indo-European), Bulgarian (South Slavic, Indo-European) and Arabic (Semitic, Non-Indo-European) \cite{campbell2013historical}. Even though Italian and Bulgarian are both Indo-European languages, Bulgarian is more distant from German than Italian in terms of Levenshtein distance \cite{Serva_2008}. By gradually adding more translated training data we monitored which additions improved the performance of the model as shown in table~\ref{tab:lang_performance}.\\
In all training settings for the fine-tuning process, training data was shuffled, weight decay was set to 0.01, batch size was 32 and improved truncation and padding was applied. The number of epochs and the learning rate were adjusted to the size of the dataset, namely the amount of sentences. All experiments were conducted on a remote server equipped with an
NVIDIA Tesla T4 GPU with 66 GB of RAM. Tokenizer and dataloader used for testing were the same ones that were used during training. The different fine-tuned models were compared regarding their macro F1-scores. The addition of the English, Italian, and Bulgarian translated training data increased the F1-score, but when including the Arabic data, it dropped. The F1-scores of the de-en-it-model and de-en-it-bg-model were very close, so we ran those fine-tuned models also on the dev-test dataset, showing a clear preference to include Bulgarian.

\begin{table*}[h]
  \caption{Training Configuration and Performance Across Languages}
  \label{tab:lang_performance}
  \begin{tabular}{cccccc}
    \toprule
    Languages & Sentences & Epochs & Learning Rate & DEV macro-F1 & DEV-TEST macro-F1 \\
    \midrule
    de & 800 & 4 & 0.001 & 0.7253 & - \\
    de, en & 1630 & 4 & 0.001 & 0.7405 & - \\
    de, en, it & 3243 & 5 & 0.001 & \textbf{0.7651} & 0.7712 \\
    de, en, it, bg & 3972 & 5 & 0.001 & \textbf{0.7692} & \textbf{0.8172} \\
    de, en, it, bg, ar & 6418 & 5 & 0.004 & 0.7275 & - \\
    \bottomrule
  \end{tabular}
\end{table*}

\paragraph{Monolingual Fine-tuning EuroBERT and XLM-RoBERTa-base}

To identify an effective architecture for monolingual subjectivity detection, we fine-tuned two transformer-based models: EuroBERT \cite{boizard2025eurobertscalingmultilingualencoders} and XLM-RoBERTa-base \cite{DBLP:journals/corr/abs-1911-02116}. 
Both models are well-suited for sentence-level classification tasks, capable of capturing nuanced semantic and syntactic patterns.
EuroBERT, a recently released multilingual model, is pre-trained primarily on European languages, aligning well with the linguistic coverage of our datasets \cite{boizard2025eurobertscalingmultilingualencoders}. 
XLM-RoBERTa-base was also selected for comparison due to its consistent cross-lingual performance and demonstrated effectiveness in prior sentence-level classification tasks \cite{DBLP:journals/corr/abs-1911-02116, StrussRuggeriBarronCedenoetal.2024}.
Experiments in monolingual setting were conducted on a remote server equipped with an NVIDIA H100 GPU with 80 GB of memory. On the provided datasets (excluding Arabic), we fine-tuned both models using a batch size of 16, a learning rate of 2e-5, and 15 training epochs.
In the mean time, we employed early stopping with a patience value of 3 based on the macro-F1 score on the development set and applied temperature scaling post-training to calibrate prediction confidence \cite{mozafari2019attendedtemperaturescalingpractical}.
We used the AdamW optimizer with a weight decay of 0.01 and employed Focal Loss \cite{lin2018focallossdenseobject} with class weighting to address label imbalance.
To improve training efficiency and stability, we adopted mixed-precision training using PyTorch’s AMP framework and applied gradient clipping with a maximum norm of 1.0 to prevent exploding gradients. After training, we evaluated both models on the dev-test datasets using a structured inference pipeline.
Input sentences were tokenized using the same configuration as during training and passed through the model to obtain raw logits.
These logits were optionally calibrated using temperature scaling and then converted to class probabilities via the softmax function.
Due to the observed class imbalance in the training data across all languages, where the SUBJ class occupies only around 37\% of instances on average and is consistently under-represented compared to OBJ, we thus applied a reduced classification threshold of 0.45 (instead of the standard 0.5) for predicting the SUBJ label.
Final predictions were mapped to their corresponding labels (\texttt{OBJ} or \texttt{SUBJ}) and compared to gold-standard labels to compute accuracy and macro-F1, as specified by the shared task organizers.


\subsection{LLM-Based Inference Strategies}
\begin{figure}[h]
    \centering
    \includegraphics[width=0.75\linewidth]{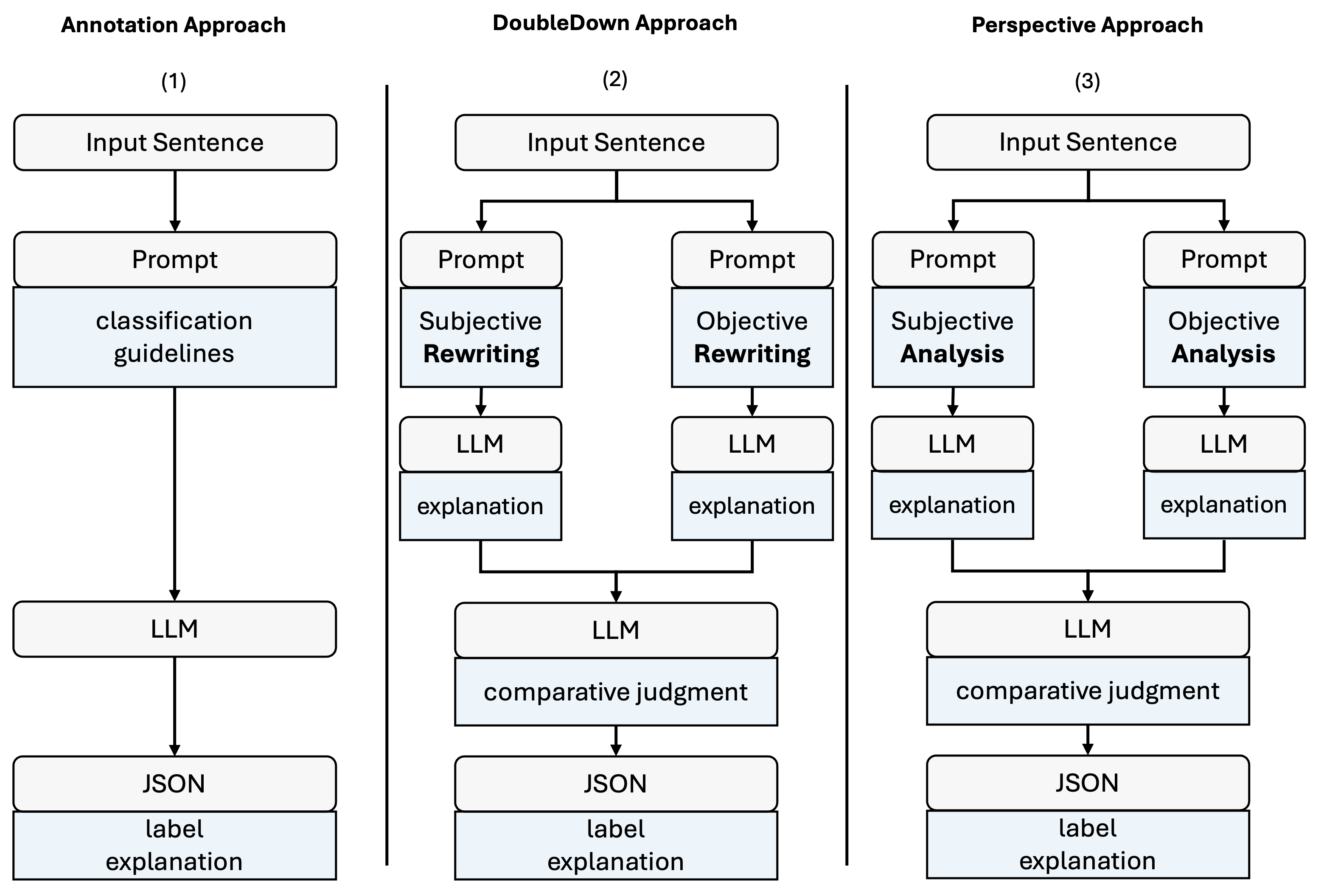}
    \caption{Overview of the three prompting strategies for subjectivity detection. In the \textbf{(1) Annotation approach}, the LLM classifies the input sentence based on explicit guidelines provided in the prompt. \textbf{(2) DoubleDown} rewrites the sentence in both subjective and objective styles and infers the label by comparing which version better preserves the original meaning. \textbf{(3) Perspective} generates explanations for both possible interpretations and selects the label based on the more convincing reasoning.}
    \label{fig:overwiew}
\end{figure}

In addition to our fine-tuned classifiers, we implement three zero-shot prompting strategies using LLMs. Each strategy frames subjectivity detection as an inference-only task and generates natural-language responses from one or more prompts per sentence. The three methods differ in reasoning style but operate directly on the input text without task-specific training, as seen in the overview in Figure \ref{fig:overwiew}.

\subsubsection{Annotation Approach}
In this inference strategy, each sentence is classified with a single API call to the OpenAI \texttt{o3-mini} model. \cite{openai2025o3mini} We supply the model with the full set of 14 decision rules \cite{antici2023corpus}, which define linguistic and rhetorical cues, to differentiate subjective from objective language. Upon receiving the model’s natural‐language response, we extract both, the verdict and the supporting explanation, in two stages: first, by parsing the response as JSON and reading the \texttt{verdict} and \texttt{explanation} fields; and, if JSON parsing fails, by performing a simple keyword match for “\textit{subjective}” or “\textit{objective}” within the raw text. In all cases, we archive both the generated explanation and the final label. The following example illustrates how each of our prompting strategies processes the same input sentence. 
\begin{tcolorbox}[colback=gray!5!white, colframe=gray!75!black]
\textbf{Input Sentence:} "Blanco established himself earlier in his career working for Dr. Luke's Kasz Money Productions."
\end{tcolorbox}

\begin{tcolorbox}[colback=gray!5!white, colframe=gray!75!black]
\textbf{Explanation:} The sentence provides factual information about Blanco's career and his affiliation with a production company. It does not include any indications of personal opinion, sarcastic remarks, or evaluative language by the author. Instead, it merely states a historical fact, which aligns with the criteria for an objective sentence. \textbf{Label:} OBJ
\end{tcolorbox}

\subsubsection{DoubleDown Approach}  
The DoubleDown approach reframes classification as a two-way comparison instead of a direct label. Rather than asking the model to judge the original sentence outright, it first generates two rewrites, one that leans into a subjective style (opinion, personal evaluation or emotional language) and a second that adopts a more objective tone (factual content, neutral phrasing). Both rewrites come from separate \texttt{gpt-4.1-mini} prompts. \cite{openai2024gpt41} In the final step, the original sentence plus both rewrites are submitted to a third prompt, which determines which version best preserves the original meaning, and on that basis classifies the input as subjective or objective. Below, you can see how the sentence is reformulated in both its subjective and objective versions.
\begin{tcolorbox}[colback=gray!5!white, colframe=gray!75!black]
\textbf{Subjective rewrite:} In my view, Blanco really made a name for himself early on thanks to his work with Dr. Luke's Kasz Money Productions collaboration that, to me, marked a crucial turning point in his career.\\
\textbf{Objective rewrite:} Blanco worked earlier in his career at Dr. Luke's Kasz Money Productions.
\end{tcolorbox}
Since the objective rewrite more closely reflects the original sentence and presents it as a factual career statement with only minor evaluative elements, the model classifies it as \textbf{OBJ}.

\subsubsection{Perspective Approach}  

The \textit{Perspective} approach frames the classification task as a stepwise evaluation of contrasting interpretations. Rather than directly predicting a label, the model is prompted to analyze the input from two distinct angles, subjective and objective, before making a decision. All prompts in this approach are handled by \texttt{gpt-4.1-mini}. \cite{openai2024gpt41} The process is divided into three steps. First, the model receives a prompt asking it to explain why the sentence might be considered subjective, focusing on aspects such as evaluative language or personal judgment. In the second step, the same sentence is analyzed from an objective perspective, highlighting features like factuality, neutrality, or verifiability. These two prompts are issued separately to ensure that the analyses are independently reasoned. In the third step, the outputs from the previous prompts are provided as input to a final classification prompt. The model is asked to compare both analyses and determine which interpretation is more convincing. It then outputs a classification along with a justification. This approach prioritizes analytical comparison over generative rewriting like in the \textit{DoubleDown} approach, as seen below. 
\begin{tcolorbox}[colback=gray!5!white, colframe=gray!75!black]
\textbf{Explanation:} The statement contains elements that can be viewed both subjectively and objectively. The subjective analysis points out that the phrase "established himself" is open to interpretation, as what qualifies as "established" can vary by individual perception, making it a somewhat evaluative judgment. The objective analysis highlights that the statement refers to a verifiable fact: Blanco worked for Dr. Luke's Kasz Money Productions earlier in his career. This part can be independently confirmed. However, the key phrase "established himself" goes beyond merely stating a fact about employment; it implies a level of success, recognition, or impact, which is inherently subjective because these concepts differ across perspectives. Therefore, while the statement contains a factual component, the primary assertion involves a subjective judgment. Given this, the subjective analysis is more convincing because the core claim revolves around the idea of "establishing oneself", which is not a strictly objective measure. \textbf{Label:} SUBJ
\end{tcolorbox}

\section{Experiments and Results}
\label{sec:experiments}

Table~\ref{tab:results} summarizes the official evaluation results of our submitted systems on the final test set, as provided by the shared task organizers. We submitted one system per subtask, using a combination of zero-shot prompting and fine-tuned transformer models. The approach and model were chosen individually for each language, based on preliminary development set results and practical observations. In the Italian monolingual subtask, our system reached the highest macro-F$_1$ score on the leaderboard (0.8104) using the \textit{Annotation Approach} with the \texttt{o3-mini} model. This suggests that prompt-based classification can work well when the input data closely follows the structure of the rules described in the prompt. For Romanian, we used a fine-tuned XLM-RoBERTa model and achieved an F1 score of 0.7917. This result placed us third overall and indicates that the model was able to generalize reasonably well, even though labeled subjectivity data in Romanian was not part of our fine-tuning. In the multilingual setting, we again used a fine-tuned \textit{XLM-RoBERTa}, which reached 0.7186 macro-F$_1$, clearly outperforming the baseline. The model showed stable results across several languages, including Greek and Polish. For German, we fine-tuned a \textit{German-BERT} model using translated training data from related languages. This approach led to solid performance (F$_1$ = 0.7269) and confirms that adding training data from similar languages can be helpful when working with limited resources, as also observed by \citeauthor{solopova2024check} for german language. In English, we used the \textit{Annotation Approach}, which achieved 0.7228 and outperformed the baseline. This result supports the idea that rule-based prompting can be effective in high-resource settings where the classification cues are well captured by the guidelines. Performance in the zero-shot subtasks varied. In Ukrainian, our system scored 0.6124, slightly below the baseline. In Greek, it reached 0.4750, showing a moderate improvement over the baseline. For Polish, the model scored 0.5665, which was slightly below the baseline (0.5719). These results suggest that zero-shot performance depends not only on the model itself, but also on the similarity between the training and test languages, and the phrasing patterns in the input data. Due to the shared task submission protocol, only one system could be submitted per language. This restriction limited our ability to systematically compare multiple approaches across all languages. Consequently, the selected system for each subtask reflects a pragmatic decision based on development performance and informal testing, rather than a globally optimal configuration. This was especially relevant for zero-shot settings, where generalization is influenced by a combination of linguistic similarity, domain coverage, and how well task framing aligns with the model's training data. Among the prompt-based methods, the Annotation strategy proved more robust than the more complex comparative prompting variants. XLM-RoBERTa consistently outperformed EuroBERT across all settings tested during development, particularly in multilingual and cross-lingual tasks.
\begin{table*}[h]
    \caption{Evaluation results for our approach across subtasks, compared to the baselines and top-ranked scores. Rows are sorted by how close our F$_1$ score is to the top result. Bold values in \textit{Our F$_1$} indicate scores above the baseline.}
  \label{tab:results}
  \begin{tabular}{lcccc}
    \toprule
    Subtask & Baseline F$_1$ & Our F$_1$ & Top Score F$_1$ & Approach \\
    \midrule
    Monolingual Italian      & 0.6941 & \textbf{0.8104} & \textbf{0.8104} & Annotation  \\
    Zero-Shot Romanian       & 0.6461 & \textbf{0.7917} & 0.8126 & XLM-RoBERTa \\
    Multilingual Subjectivity& 0.6390 & \textbf{0.7186} & 0.7550 & XLM-RoBERTa \\
    Zero-Shot Ukrainian      & 0.6296 & 0.6124 & 0.6424 & XLM-RoBERTa \\
    Zero-Shot Greek          & 0.4159 & \textbf{0.4750} & 0.5067 & XLM-RoBERTa \\
    Monolingual English      & 0.5370 & \textbf{0.7228} & 0.8052 & Annotation  \\
    Monolingual German       & 0.6960 & \textbf{0.7269} & 0.8520 & German-BERT \\
    Zero-Shot Polish         & 0.5719 & 0.5665 & 0.6922 & XLM-RoBERTa \\
    \bottomrule
  \end{tabular}
\end{table*}


\section{Conclusion and Future Work}
\label{sec:conclusion}
We presented a multilingual system for subjectivity detection using two main approaches: fine-tuned transformer models and zero-shot prompting with LLMs. Our results in the CheckThat! 2025 shared task show that both directions can be effective, depending on language and resource availability. For fine-tuned models, XLM-RoBERTa delivered the most consistent performance and was used in several subtasks, including multilingual and zero-shot settings. For German, we observed that fine-tuning a German-BERT model with translated training data led to competitive results. On the prompting side, the Annotation Approach with the o3-mini model performed well in high-resource languages, such as Italian and English, where classification rules were clearly reflected in the data. Due to the submission constraint of only one system per language, we could not test all combinations of models and approaches systematically. Our choices were based on limited development set results and informal comparisons. This affected our ability to fully explore the strengths and weaknesses of each approach across languages, especially for zero-shot cases. Another important limitation was the lack of broader context for each sentence. Since the task involved classifying isolated sentences, it was often difficult to judge subjectivity accurately without context. This made the task especially challenging when subjective language relied on surrounding sentences. Another issue is the imbalance of label distributions across languages, that is most notably in the Bulgarian development set, where subjective sentences dominate. These imbalances can lead models to internalize and amplify misleading associations, potentially reinforcing biases and over-predicting subjectivity in certain languages or cultural contexts. For future work, it would be valuable to systematically compare prompting and fine-tuned approaches across languages and subtasks under controlled conditions. In particular, we aim to better understand which types of tasks or linguistic features favor instruction-based inference over supervised training. Additionally, exploring more flexible combinations of prompting and fine-tuning, e.g., via model ensembling or fallback strategies such as few-shot prompting or confidence-based model switching, could help improve performance, especially in low-resource or zero-shot settings.

\begin{acknowledgments} 
This research is funded by the Federal Ministry of Research, Technology and Space (BMFTR, reference: 03RU2U151C) in the scope of the research project news-polygraph.
\end{acknowledgments}



\bibliography{sample-ceur}

\appendix



\end{document}